\documentclass{article}
\usepackage{spconf,amsmath}
\usepackage{url}
\usepackage{cite}
\usepackage{xcolor}
\usepackage{bm}
\usepackage{times}
\usepackage{graphicx}
\usepackage{amssymb}
\usepackage{bm}

\title{Variational Inference for Background Subtraction in Infrared Imagery}
\name{K. Makantasis$^1$, A. Doulamis$^2$, N. Doulamis$^2$\thanks{}}
\address{$^1$Technical University of Crete, University campus, Kounoupidiana, 73100, Chania, Greece \\ $^2$National Technical University of Athens, Zographou campus, 15780, Athens, Greece\\
{\normalsize  \url{kmakantasis@isc.tuc.gr}; \url{adoulam@cs.ntua.gr}; \url{ndoulam@cs.ntua.gr}}
}
\usepackage{setspace}
\begin{document}
\onecolumn
%
\maketitle

\begin{abstract}
We propose a Gaussian mixture model for background subtraction in infrared imagery. Following a Bayesian approach, our method automatically estimates the number of Gaussian components as well as their parameters, while simultaneously it avoids over/under fitting. The equations for estimating model parameters are analytically derived and thus our method does not require any sampling algorithm that is computationally and memory inefficient. The pixel density estimate is followed by an efficient and highly accurate updating mechanism, which permits our system to be automatically adapted to dynamically changing operation conditions. Experimental results and comparisons with other methods show that our method outperforms, in terms of precision and recall, while at the same time it keeps computational cost suitable for real-time applications.  
\end{abstract}

\section{Introduction}

Pixels values of infrared frames correspond to the relative differences in the amount of thermal energy emitted or reflected from objects in the scene. Due to this fact, infrared cameras are equally applicable for both day and night scenarios, while at the same time, compared to visual-optical cameras, are less affected by changing illumination or background texture. Furthermore, infrared imagery eliminates any privacy issues as people being depicted in the scene can not be identified \cite{gade_long-term_2013}. These features make infrared cameras prime candidate for persistent video surveillance systems.

\begin{figure}[t]
\begin{minipage}[b]{1.0\linewidth}
  \centering
  \centerline{\includegraphics[width=13.5cm]{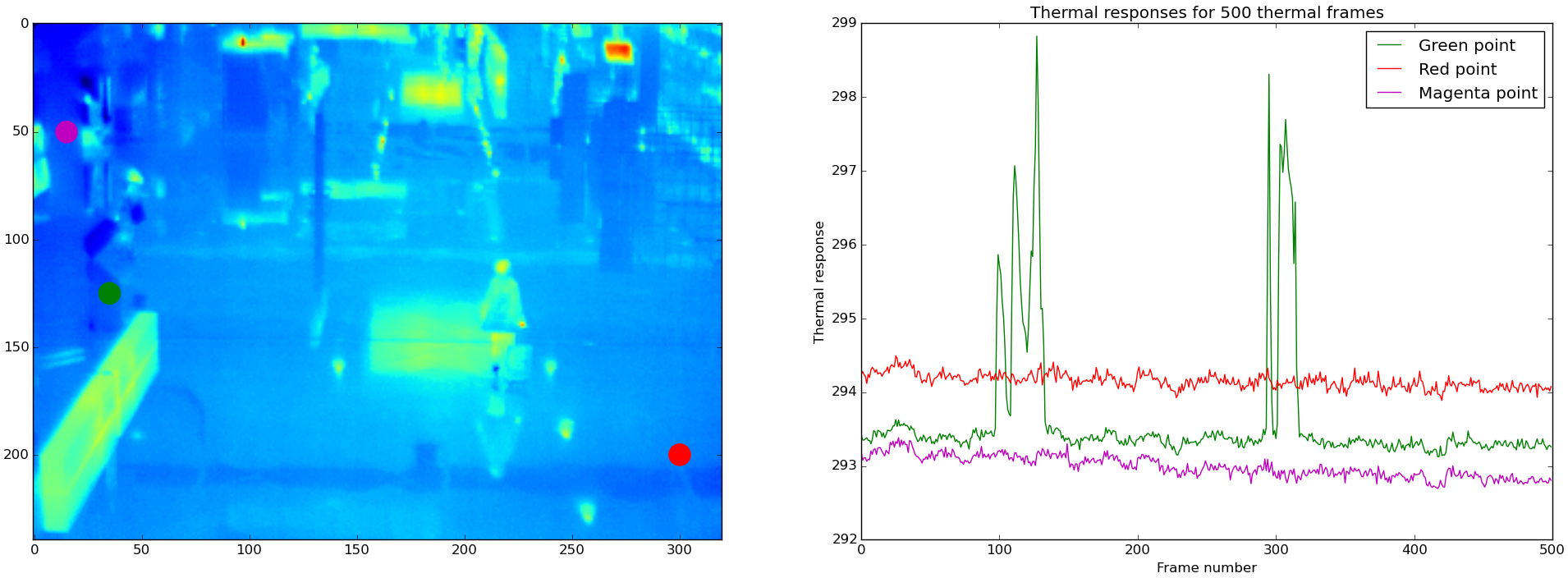}}
\end{minipage}
\caption{Thermal responses for three different points. In contrast to visual-optical videos, where pixels take integer values, thermal responses are floating point numbers, corresponding to objects' temperature.} 
\label{fig:Thermal responses}
\end{figure}

Although, infrared imagery can alleviate several problems associated with visual-optical videos, it has its own unique challenges such as a) low signal-to-noise ratio (noisy data) and b) almost continuous pixel values that model objects' temperature. Both issues complicate pixel responses modeling. An example of raw thermal responses is presented in Figure \ref{fig:Thermal responses}, where pixel values are floating point numbers ranging from 293 to 299 Kelvin degrees. Due to this peculiarity most of conventional computer vision techniques, that successfully used for visual-optical data, can not be applied straightforward on thermal imagery.

For many high-level vision based applications, either they use visual-optical videos \cite{cheung_robust_2005, porikli_achieving_2006, tuzel_human_2007, tuzel_pedestrian_2008} or infrared data \cite{jungling_feature_2009, latecki_tracking_2005, wang_improved_2010}, the task of background subtraction constitutes a key component, as this is one of the most common methods for locating moving objects, facilitating search space reduction and visual attention modeling.

Background subtraction techniques applied on visual-optical videos model the color properties of depicted objects \cite{brutzer_evaluation_2011, herrero_background_2009} and can be classified into three main categories \cite{el_baf_fuzzy_2009}: basic background modeling \cite{mcfarlane_segmentation_1995, zheng_extracting_2006}, statistical background modeling \cite{elgammal_non-parametric_2000, wren_pfinder:_1997} and background estimation \cite{messelodi_kalman_2005, toyama_wallflower:_1999}. The most used methods are the statistical ones due to their robustness to critical situations. In order to statistically represent the background, a probability distribution is used to model the history of pixel values intensity over time. Towards this direction, the work of Stauffer and Grimson \cite{stauffer_adaptive_1999}, is one of the best known approaches. It uses a Gaussian mixture model, with fixed number of components, for a per-pixel density estimate. Similar to this approach, Makantasis \textit{et al.} in \cite{makantasis_student-t_2012} propose a Student-t mixture model for background modeling, taking advantage of Student-t distribution compactness and robustness to noise and outliers. The works of \cite{zivkovic_improved_2004} and \cite{zivkovic_efficient_2006} extend the method of \cite{stauffer_adaptive_1999} by introducing a rule based on a user defined threshold to estimate the number of components. However, this rule is application dependent and not directly derived from the data. Haines and Xiang in \cite{haines_background_2014} address this drawback by using a Dirichlet process mixture model. Due to the computational cost of their method, the authors propose a GPU implementation. All of the aforementioned techniques present the drawback that objects' color properties are highly affected by scene illumination, making the same object to look completely different under different lighting or weather conditions.
 
Although, thermal imagery can provide a challenging alternative for addressing the aforementioned difficulty, there exist few works for thermal data. The authors of \cite{davis_fusion-based_2005, davis_robust_2004, davis_background-subtraction_2007} exploit contour saliency to extract foreground objects. Initially, they utilize a unimodal background modeling technique to detect regions of interest and then exploit the halo effect of thermal data for extracting foreground objects. However unimodal background modeling is not usually capable of capturing background dynamics. Baf \textit{et al.} in \cite{el_baf_fuzzy_2009} present a fuzzy statistical method for background subtraction to incorporate uncertainty into the mixture of Gaussians. However, this method requires a predefined number of components making this approach to be application dependent. Elguebaly and Bouguila in \cite{elguebaly_finite_2013} propose a finite asymmetric generalized Gaussian mixture model for object detection. Again this method requires a predefined maximum number of components, presenting therefore limitations when this technique is applied on uncontrolled environments. Dai \textit{et al.} in \cite{dai_pedestrian_2007} propose a method for pedestrian detection and tracking using infrared imagery. This method consists of a background subtraction technique that exploits a two-layer representation (one for foreground and one for background) of infrared frame sequences. However, the assumption made is that the foreground is restricted to moving objects, a consideration which is not sufficient for dynamically changing environments. One way to handle the aforementioned difficulties is to introduce a background model, the parameters and the structure of which are directly estimated from the data, while at the same time it takes into account the specific properties of infrared imagery. 

\subsection{Our contribution}
\label{sec:related work}

This work presents background modeling able to provide a per pixel density estimate, taking into account the special characteristics of infrared imagery, such as low signal-to-noise ratio. Our method exploits a Gaussian mixture model with unknown number of components. The advantage of such a model is that its own parameters and structure can be directly estimated from the data, allowing dynamic model adaptation to uncontrolled and changing environments.

An important issue in the proposed Gaussian mixture modeling concerns learning the model parameters. In our method, this is addressed using a variational inference framework to correspond the functional structure of the model with real data distributions obtained from the infrared images. Then, the Expectation-Maximization (EM) algorithm is adopted to fit the outcome of variational inference to real measurements. Updating procedures are incorporated to allow dynamic model adaptation to the forthcoming infrared data. Our updating method avoids of using heuristics by considering existing knowledge accumulated from previous data distributions and then it compensates this knowledge with current measurements.      

Our overall strategy exploits a Bayesian framework to estimate all the parameters of the  mixture model and thus avoiding over/under fitting issues. To compensate computational challenges arising from the non a priori known nature of the mixture model, we utilize conjugate priors and thus we derive analytical equations for model estimation. In this way, we avoid the need of any sampling method, which are computationally and memory inefficient. This selection makes our system suitable for online and real-time applications.

This paper is organized as follows: Section \ref{sec:problem formulation} formulates the Bayesian framework for mixture modeling. In Section \ref{sec:derivation of random variables distribution} we present the procedure to analytically derive the solutions for estimating the distributions of model parameters. Section \ref{sec:random variables optimization} describes the EM algorithm along with setting the priors and parameters initialization. Section \ref{sec:online adaptation mechanism} discusses the online updating mechanism and Section \ref{sec:background subtraction} presents the background subtraction task. In Section \ref{sec:experimental results} experimental results are given and Section \ref{sec:conclusions} concludes this work.

\section{Gaussian mixture modeling}
\label{sec:problem formulation}

In this section we formulate the Bayesian framework adopted in this paper to analytically estimate all the parameters of the proposed Gaussian mixture model. For this reason, in section \ref{sec:Infinite Gaussian mixture model} we briefly describe the basic theory behind Gaussian mixture model, while in section \ref{sec:conjugate priors} we describe the introduction of conjugate priors that assist us in yielding analytical model estimations as in Section \ref{sec:derivation of random variables distribution}.

\subsection{Model fundamentals}
\label{sec:Infinite Gaussian mixture model}

The Gaussian mixture distribution can be seen as a linear superposition of Gaussian functional components,
\begin{equation}
p(x|\bm \varpi, \bm \mu, \bm \tau) = \sum_{k=1}^K \varpi_k \mathcal{N}(x|\mu_k, \tau_k^{-1})
\label{eq:gaussian_mm}
\end{equation}
where the parameters $\{\varpi_k\}_{k=1}^K$ must satisfy $0 \leq \varpi_k \leq 1$ for every $k$ and $\sum_{k=1}^K \varpi_k = 1$ and $K$ is the number of Gaussian components. In the proposed mixture modeling, variable $K$ can take any natural value up to infinity. However, it is highly recommended to set the upper bound for $K$ less than the cardinality of the dataset, i.e. the number of observed samples. By introducing a $K$-dimensional binary latent variable $\bm z$, such as $\sum_{k=1}^K z_k= 1$ and $p(z_k=1) = \varpi_k$, the distribution $p(x)$ can be defined in terms of a marginal distribution $p(\bm z)$ and a conditional distribution $p(x|\bm z)$ as follows
\begin{equation}
p(x|\bm \varpi, \bm \mu, \bm \tau) = \sum_{\bm z} p(\bm z | \bm \varpi) p(x| \bm z, \bm \mu, \bm \tau)
\label{eq:gaussian_mm_z}
\end{equation}
where $p(\bm z|\bm \varpi)$ and and $p(x|\bm z)$ are in the form of
\begin{equation}
p(\bm z | \bm \varpi) = \prod_{k=1}^{K} \varpi_k^{z_k}
\label{eq:p_z}
\end{equation}
\begin{equation}
p(x|\bm z, \bm \mu, \bm \tau) = \prod_{k=1}^{K} \mathcal{N}(x|\mu_k, \tau_k^{-1})^{z_k}
\label{eq:p_x}
\end{equation}
where $\bm \mu = \{\mu_k\}_{k=1}^K$ and $\bm \tau = \{\tau_k\}_{k=1}^K$, correspond to the mean values and precisions of Gaussian components. By introducing latent variables and transforming the Gaussian mixture distribution into the form of (\ref{eq:gaussian_mm_z}), we are able to exploit the EM algorithm for fitting our model to the observed data, as shown in Section \ref{sec:random variables optimization}. 

If we have in our disposal a set $\bm X = \{x_1,...,x_N\}$ of observed data we will also have a set $\bm Z = \{\bm z_1,...,\bm z_N \}$ of latent variables. Each $\bm z_n$ will be a $K$-dimensional binary vector, such as $\sum_{k=1}^K z_{nk}= 1$, and, in order to take into consideration the whole dataset of $N$ samples, the distributions of (\ref{eq:p_z}) and (\ref{eq:p_x}) will be transformed to
\begin{equation}
p(\bm Z | \bm \varpi) = \prod_{n=1}^{N} \prod_{k=1}^{K} \varpi_k^{z_{nk}}
\label{eq:p_Z}
\end{equation}
\begin{equation}
p(\bm X|\bm Z, \bm \mu, \bm \tau) = \prod_{n=1}^{N} \prod_{k=1}^{K} \mathcal{N}(x_n|\mu_k, \tau_k^{-1})^{z_{nk}}
\label{eq:p_X}
\end{equation}

\subsection{Conjugate priors}
\label{sec:conjugate priors}

To avoid computational problems in estimating the parameters and the structure of the proposed Gaussian model, we introduce conjugate priors, over the model parameters $\bm \mu$, $\bm \tau$ and $\bm \varpi$,  that allow us to yield analytical solutions. This way the need of using sampling methods is prevented. Introduction of priors implies the use of a Bayesian framework for the analysis.

Let us denote as $\bm Y=\{\bm Z, \bm \varpi, \bm \mu, \bm \tau\}$ the set which contains all model latent variables and parameters and as $q(\bm Y)$ its distribution. Then, our goal is to estimate $q(\bm Y)$ which maximizes model evidence $p(\bm X)$.

\begin{equation}
q(\bm Y) : \max \ln p(\bm X)
\label{eq:maximize_px}
\end{equation}
where in (\ref{eq:maximize_px}) we used the logarithm of $p(\bm X)$ for calculus purposes. For maximizing (\ref{eq:maximize_px}) we need to define the distribution over $\bm Y$, that is, $p(\bm Z|\bm \varpi)$ from (\ref{eq:p_Z}), $p(\bm \varpi)$ and $p(\bm \mu, \bm \tau)$.
 
Due to the fact that $p(\bm Z | \bm \varpi)$ is a Multinomial distribution, its conjugate prior is a Dirichlet distribution over the mixing coefficients $\bm \varpi$
\begin{equation}
p(\bm \varpi) = \frac{\Gamma(K\lambda_0)}{\Gamma(\lambda_0)^K} \prod_{k=1}^{K}\varpi_k^{\lambda_0-1}
\label{eq:varpi_prior}
\end{equation} 
where $\Gamma(\cdot)$ is the Gamma function. Parameter $\lambda_0$ has a physical interpretation; the smaller the value of this parameter is, the larger is the influence of the data rather than the prior on the posterior distribution $p(\bm Z|\bm \varpi)$. In order to introduce uninformative priors and not prefer a specific component against the other, we choose to use a single parameter $\lambda_0$ for the Dirichlet distribution, instead of a vector with different values for each mixing coefficient. 

Similarly, the conjugate prior of (\ref{eq:p_x}) takes the form of a Gaussian-Gamma distribution, because both $\bm \mu$ and $\bm \tau$ are unknown. Subsequently, the joint distribution of parameters $\bm \mu$ and $\bm \tau$ can be modeled as 
\begin{subequations}
\begin{align}
p(\bm \mu, \bm \tau & )  = p(\bm \mu | \bm \tau)p(\bm \tau) \\
& = \prod_{k=1}^{K} \mathcal{N}(\mu_k|m_0,(\beta_0\tau_k)^{-1})Gam(\tau_k|a_0,b_0)
\end{align}
\label{eq:joint_mu_tau_prior}
\end{subequations}
where $Gam(\cdot)$ denotes the Gamma distribution.

In order to not express any specific preference about the form of the Gaussian components through the introduction of priors, we use uninformative priors by setting the values of hyperparameters $m_0$, $\beta_0$, $a_0$ and $b_0$ to appropriate values as shown in Section \ref{sec:random variables optimization}.

Having defined the parametric form of observed data, latent variables and parameters distributions, our goal is to approximate the posterior distribution $p(\bm Y| \bm X)$ and the model evidence $p(\bm X)$, where $\bm Y=\{\bm Z, \bm \varpi, \bm \mu, \bm \tau\}$ is the set with distribution $q(\bm Y)$, which contains all model latent variables and parameters. Based on the Bayes rule, which states that $p(\bm X) p(\bm Y|\bm X) = p(\bm X, \bm Y)$ the logarithm of distribution $p(\bm X)$ can be expressed as 
\begin{subequations}
\begin{align}
\ln p(\bm X) & = \int q(\bm Y) \ln \frac{p(\bm X, \bm Y)}{q(\bm Y)} d\bm Y - \\ \nonumber
&- \int q(\bm Y) \ln \frac{p(\bm Y | \bm X)}{q(\bm Y)} d\bm Y \\
& = \mathcal{L}(q) + KL(q||p)
\end{align}
\label{eq:log_px}
\end{subequations}
where $KL(q||p)$ is the Kullback-Leibler divergence between $q(\bm Y)$ and $p(\bm Y | \bm X)$ distributions and $\mathcal{L}(q)$ is the lower bound of $\ln p(\bm X)$. 
Since $KL(q||p)$ is a non negative quantity, equals to zero only if $q(\bm Y)$ is equal to $p(\bm Y | \bm X)$, maximization of $\ln p(\bm X)$ is equivalent to minimizing of $KL(q||p)$. For minimizing $KL(q||p)$ and estimating $p(\bm X)$ we exploit the EM algorithm, as shown in Section \ref{sec:random variables optimization}.

By making the assumption, based on variational inference, that the distribution $q(\bm Y)$ can be factorized over $M$ disjoint sets such as $q(\bm Y) = \prod_{i=1}^{M}q_i(\bm Y_i)$,
as shown in \cite{bishop_pattern_2007}, the optimal solution $q_j^*(Y_j)$ corresponds to the minimization of $KL(q||p)$ is given by
\begin{equation}
\ln q_j^*(\bm Y_j) = \mathbb{E}_{i \neq j} [ \ln p(\bm X, \bm Y) ] + \mathcal{C}
\label{eq:q_star}
\end{equation}
where $\mathbb{E}_{i \neq j} [ \ln p(\bm X, \bm Y) ]$ is the expectation of the joint distribution over all variables $\bm Y_j$ for $j \neq i$ and $\mathcal{C}$ is a constant. $P(\bm X, \bm Y)$ is modeled through (\ref{eq:joint_factorization}). 

In the following, we present the analytical solution for the optimal distributions $q_j^*(Y_j)$ for model parameters and latent variables, i.e. the optimal distributions $q^*(\bm Z)$, $q^*(\bm \varpi)$, $q^*(\bm \tau)$ and $q^*(\bm \mu|\bm \tau)$.

\section{Optimal model parameter distributions}
\label{sec:derivation of random variables distribution}
According to (\ref{eq:p_Z}), (\ref{eq:p_X}), (\ref{eq:varpi_prior}) and (\ref{eq:joint_mu_tau_prior}), the joint distribution of all random variables can be factorized as follows
\begin{equation}
\begin{aligned}
p(\bm X, \bm Z, \bm \varpi, \bm \mu, \bm \tau) = & p(\bm X|\bm Z, \bm \mu, \bm \tau) p(\bm Z|\bm \varpi) \\  
& p(\bm \varpi) p(\bm \mu|\bm \tau) p(\bm \tau)
\end{aligned}
\label{eq:joint_factorization}
\end{equation}
$\bm X$ corresponds to the set of the observed variables. All proofs are given in Appendix \ref{ap:appendix}.

\subsection{Optimal $q^*(Z)$ distribution}
Using (\ref{eq:q_star}) and the factorized form of (\ref{eq:joint_factorization}) the distribution of the optimized factor $q^*(\bm Z)$ is given by a Multinomial distribution of the form
\begin{subequations}
\begin{align}
& q^*(\bm Z) = \prod_{n=1}^{N}\prod_{k=1}^{K}\bigg(\frac{\rho_{nk}}{\sum_{j=1}^{K}\rho_{nj}}\bigg)^{z_{nk}} = \\ 
& \:\:\:\:\:\:\:\:\:\:\:\:= \prod_{n=1}^{N}\prod_{k=1}^{K} r_{nk}^{z_{nk}}
\label{eq:q_Z_optimized_dirichlet_b}
\end{align}
\label{eq:q_Z_optimized_dirichlet}
\end{subequations}  
as $\rho_{nk}$ we have denote the quantity
\begin{equation}
\begin{aligned}
\rho_{nk} = \exp\bigg(& \mathbb{E}\big[\ln \varpi_k \big] + \frac{1}{2}\mathbb{E}\big[\ln \tau_k\big] -\frac{1}{2}\ln2\pi - \\ &-\frac{1}{2} \mathbb{E}_{\bm \mu, \bm \tau}\big[(x_n-\mu_k)^2\tau_k \big]\bigg)
\end{aligned}
\label{eq:pho_nk}
\end{equation}
The expected value $\mathbb{E}[z_{nk}]$ of $q^*(\bm Z)$ is equal to  $r_{nk} $.


\subsection{Optimal $q^*(\varpi)$ distribution}
Using (\ref{eq:joint_factorization}) and (\ref{eq:q_star}) the distribution of the optimized factor $q^*(\bm \varpi)$ is given a Dirichlet distribution of the form 
\begin{equation}
q^*(\bm \varpi) = \frac{\Gamma(\sum_{i=1}^{K}\lambda_i)}{\prod_{j=1}^{K}\Gamma(\lambda_j)} \prod_{k=1}^{K}\varpi_k^{\lambda_k - 1}
\label{eq:q_star_varpi_derivation}
\end{equation}
$\lambda_k$ is equal to $N_k + \lambda_0$, where $N_k=\sum_{n=1}^{N}r_{nk}$ represents the proportion of data that belong to the $k$-th component.

\subsection{Optimal $q^*(\mu_k | \tau_k)$ distribution}
Similarly, the distribution of the optimized factor $q^*(\mu_k, \tau_k)$ is given by a Gaussian distribution of the form 
\begin{equation}
q^*(\mu_k|\tau_k) = \mathcal{N}(\mu_k|m_k, (\beta_k \tau)^{-1})
\label{eq:q_star_mu_distribution}
\end{equation}
where the parameters $m_k$ and $\beta_k$ are given by 
\begin{subequations}
\begin{align}
\beta_k & = \beta_0 + N_k \\
m_k & = \frac{1}{\beta_k}\Big(\beta_0 m_0 + N_k \bar x_k\Big)
\end{align}
\label{eq:q_star_mk_tauk}
\end{subequations}
where $\bar x_k$ is equal to $\frac{1}{N_k}\sum_{n=1}^{N}r_{nk}x_n$ represents the centroid of the data that belong to the $k$-th component.

\subsection{Optimal $q^*(\tau_k)$ distribution}
After the estimation of $q^*(\mu_k|\tau_k)$, distribution of the optimized factor $q^*(\tau_k)$ is given by a Gamma distribution of the following form 
\begin{equation}
q^*(\tau_k) = Gam(\tau_k|a_k, b_k)
\label{eq:q_star_tau_distribution}
\end{equation}
while the parameters $a_k$ and $b_k$ are given by the following relations 
\begin{subequations}
\begin{align}
a_k & = a_0 +  \frac{N_k}{2} 
\label{eq:q_star_tauk_a}\\
b_k & = b_0 + \frac{1}{2}\bigg(N_k\sigma_k + \frac{\beta_0 N_k}{\beta_0 + N_k}\big(\bar x_k - m_0\big)^2 \bigg)
\label{eq:q_star_tauk_b}
\end{align}
\label{eq:q_star_tauk}
\end{subequations}
where $\sigma_k = \frac{1}{N_k}\sum_{n=1}^{N}(x_n-\bar x_k)^2$.

\section{Distribution parameters optimization}
\label{sec:random variables optimization}

After the approximation of random variables distributions, we will use the EM algorithm in order to find optimal values for model parameters, i.e. maximize (\ref{eq:log_px}). In order to use the EM algorithm, we have to initialize priors hyperparameters $\lambda_0$, $a_0$, $b_0$, $m_0$ and $\beta_0$ and the model parameters $\varpi_k$, $\mu_k$, $\tau_k$, $\beta_k$, $a_k$, $b_k$ and $\lambda_k$ (see Section \ref{sec:derivation of random variables distribution}).

The parameter $\lambda_0$ can be interpreted as the effective prior number of observations associated with each component. In order to introduce an uninformative prior for $\bm \varpi$, we set the parameter $\lambda_0$ equal to $N/K$, suggesting that the same number of observations is associated to each component. Parameters $a_0$ and $b_0$ (positive values due to Gamma distribution) were set to the value of $10^{-3}$. Our choice is justified by the fact that the results of updating equations (\ref{eq:q_star_tauk_a}) and (\ref{eq:q_star_tauk_b}) are primarily affected by the data and not by the prior when the values for $a_0$ and $b_0$ are close to zero. The mean values of the components are described by conditional Normal distribution with means $m_0$ and precisions $\beta_0 \tau_k$. We introduce an uninformative prior by setting the value for $m_0$ to the mean of the observed data and the parameter $\beta_0=\frac{b_0}{a_0 v_0}$, where $v_0$ is the variance of the observed data.

The convergence of EM algorithm is facilitated by initializing the parameters $\varpi_k$, $\mu_k$, $\tau_k$ and $\beta_k$ using the k-means. To utilize k-means, the number of clusters, corresponding to the Gaussian components, should be a priori known. Since we create a mixture model, the number of Gaussian components should be less or equal to the number of observed data. For this reason we set the number of clusters $K_{max}$ to a value smaller or equal to the number of observations. If we have no clue about the number of classes we can set $K_{max}$ to equal $N$. If we denote as $\hat N_k$ the number of observation that belong to $k$-th cluster, then we can set the value of parameter $\mu_k$ to equal the centroid of $k$-th cluster, the parameter $\varpi_k$ to equal the proportion of observations for the $k$-th cluster, the parameter $\tau_k$ to equal $\hat v_k^{-1}$, where $v_k$ stands for the variance of the data of the $k$-th cluster and the parameter $\beta_k$ to equal $\hat N_k^{-1}$. Having initialized the parameters $\varpi_k$, $\mu_k$, $\tau_k$ and $\beta_k$, we can exploit the formula for the expected value of a Gamma distribution to initialize the parameters $a_k$ and $b_k$ to values $\tau_k$ and one respectively. Finally, the initialization of $\varpi_k$ allows us to initialize the parameter $\lambda_k$, which can be interpreted as the effective number of observations associated with each Gaussian component, to the value $N\varpi_k$.  

After the initialization of model parameters and priors hyperparameters, the EM algorithm can be used to minimize $KL(q||p)$ of (\ref{eq:log_px}). During the E step, $r_{nk}$ is calculated given the initial/current values of all the parameters of the model. Using (\ref{eq:q_Z_optimized_dirichlet_b}) $r_{nk}$ is given by

\begin{equation}
r_{nk} \propto \tilde{\varpi_k} \tilde{\tau_k}^{1/2} \exp \bigg(-\frac{a_k}{2b_k}\big(x_n-m_k\big)^2 - \frac{1}{2\beta_k} \bigg) 
\label{eq:r_nk}
\end{equation}
Due to the fact that $q^*(\bm \varpi)$ is a Dirichlet distribution and $q^*(\tau_k)$ is a Gamma distribution, $\tilde{\varpi_k}$ and $\tilde{\tau_k}$ will be given by
\begin{subequations}
\begin{align}
& \ln \tilde{\varpi_k} \equiv \mathbb{E}\big[\ln \varpi_k\big] = \Psi(\lambda_k) - \Psi\bigg(\sum_{k=1}^{K} \lambda_k\bigg) \\
& \ln \tilde{\tau_k} \equiv \mathbb{E}\big[\ln \tau_k\big] = \Psi(a_k) - \ln b_k
\end{align}
\label{eq:tilde_pi_tilde_tau}
\end{subequations}
where $\Psi(\cdot)$ is the digamma function.

During the M step, we keep fixed the value for variables $r_{nk}$ (the value that was calculated during the E step), and we re-calculate the values for model parameters using (\ref{eq:q_star_varpi_derivation}), (\ref{eq:q_star_mk_tauk}) and (\ref{eq:q_star_tauk}). The steps E and M are repeated sequentially untill the values for model parameters are not changing anymore.  As shown in \cite{boyd_convex_2004} convergence of EM algorithm is guaranteed because bound is convex with respect to each of the factors $q(\bm Z)$, $q(\bm \varpi)$, $q(\bm \mu|\bm \tau)$ and $q(\bm \tau)$. 

During model training the mixing coefficient for some of the components takes value very close to zero. Components with mixing coefficient less than $1/N$ are removed (we require each component to model at least one observed sample) and thus after training the model has automatically determined the right number of Gaussian components.

\section{Online updating mechanism}  
\label{sec:online adaptation mechanism}
Having described how our model fits to $N$ observed data, in this section we present the mechanism that permits our model to automatically adapt to new observed data. We use no heuristic rules but statistics based on the observed data.

Let us denote as $x_{new}$ a new observed sample. Then, there are two cases; either the new observed sample is successfully modeled by our trained model, or not. To estimate if a new sample is successfully modeled, we find the closest component to the new sample. As a distance metric between components and the new sample, we use the Mahalanobis distance, since this is reliable distance  measure between a point and a distribution. 

The closest component $c$ to the new sample is the one that presents the minimum Mahalanobis distance $D_k$ 
\begin{equation}
c = \arg \min_k D_k = \arg \min_k \sqrt{(x_{new}-\mu_k)^2 \tau_k}
\label{eq:Mahalanobis}
\end{equation}
The probability of the new sample to belong to $c$
\begin{equation}
p(x_{new}|\mu_c, \tau_c) = \mathcal{N}(x_{new}|\mu_c, \tau_c^{-1})
\label{eq:p_mix}
\end{equation}
where $\mu_c$ and $\tau_c$ stand for the closest component mean value and precision respectively.

Let us denote as $\Omega$ the initially observed dataset. Then, we can assume that the probability to observe the new sample $x_{new}$ is given by
\begin{equation}
p(x_{new}|e) = \frac{N_e}{N} \mathcal{U}(x_{new}|x_{new}-e, x_{new}+e)
\label{eq:p_xnew}
\end{equation}
where $N_e=\big|\{ x_i \in \Omega : x_{new}-e \leq x_i \leq x_{new}+e \}\big|$ and $\mathcal{U}(x_{new}|x_{new}-e, x_{new}+e)$ is a Uniform distribution with lower and upper bounds to equal $x_{new}-e$ and $x_{new}+e$ respectively. Equation (\ref{eq:p_xnew}) suggests that the probability to observe $x_{new}$ is related to the proportion of data that have already been observed around $x_{new}$. By increasing the neighborhood around $x_{new}$, i.e. increasing the value of $e$, the quantity $\mathcal{U}(x_{new}|x_{new}-e, x_{new}+e)$ is decreasing, while the value of $N_e$ is increasing. 

Upon arrival of a new sample $x_{new}$, we can estimate the optimal range $\epsilon$  around $x_{new}$ that maximizes (\ref{eq:p_xnew}) as
\begin{equation}
\epsilon = \arg \max_e p(x_{new}|e)
\label{eq:epsilon}
\end{equation}

Then, if $p(x_{new}|\mu_c, \tau_c) \geq p(x_{new}|\epsilon)$ the new observed sample $x_{new}$ can sufficiently represented by our model. Otherwise, a new Gaussian component must be created.

\begin{figure}[t]
\begin{minipage}[b]{1.0\linewidth}
  \centering
  \centerline{\includegraphics[width=12.0cm]{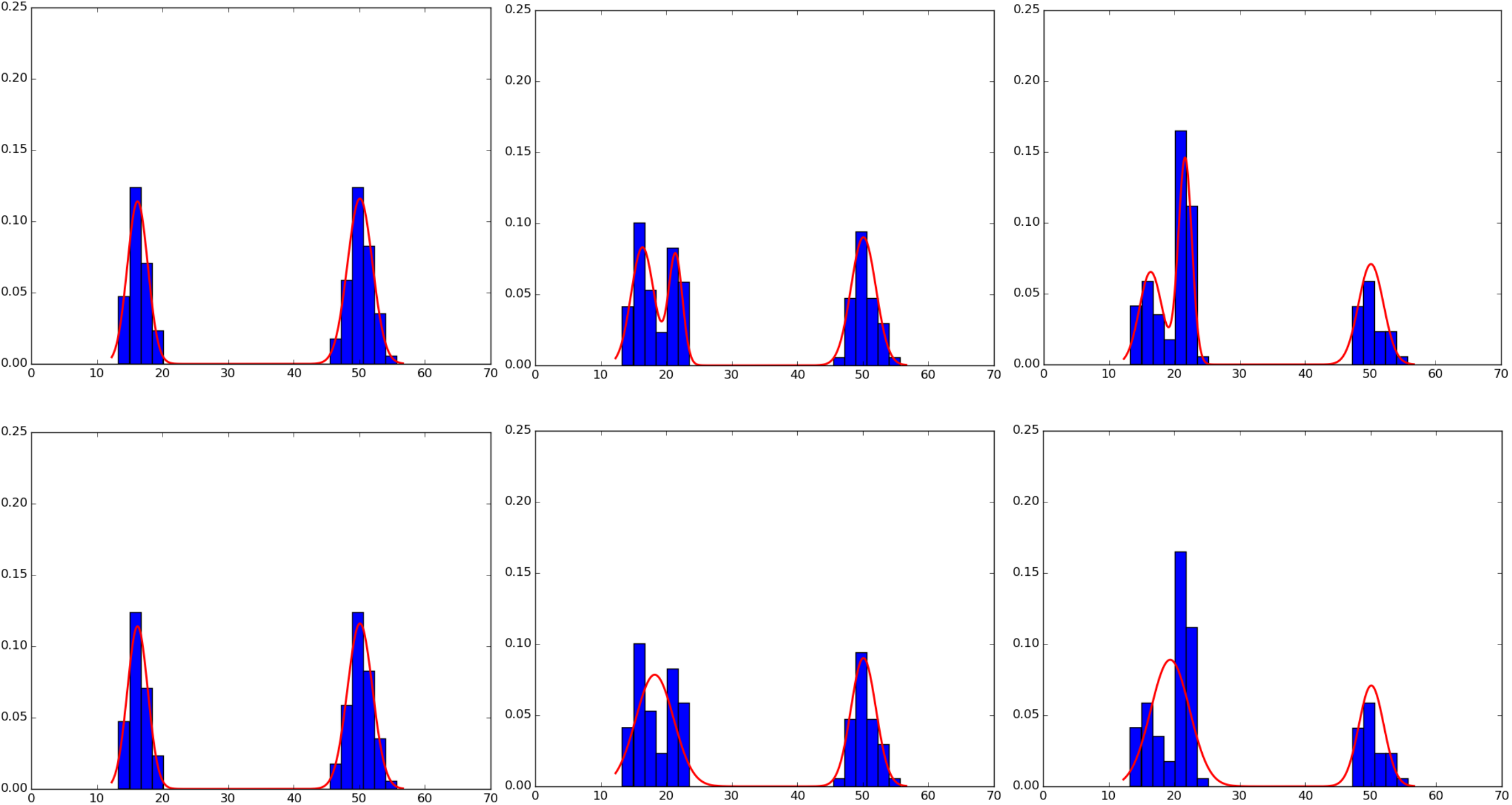}}
\end{minipage}
\caption{First row: updating of our model to new observed data. Second row: updating of model presented in \cite{zivkovic_improved_2004}.}
\label{fig:km vs mog}
\end{figure} 

For model updating, we need to define the binary variable $o$, called ownership and associated with the Gaussian components, as
\begin{equation}
o_k = \begin{cases} 1, & \mbox{if } k=c \\ 0, & \mbox{otherwise} \end{cases}
\end{equation}
where we recall that $c$ represents the index of the closest component and $k$ is the index of $k$-th component.

When the new observed sample is successfully modeled, the parameters for the Gaussian components are updated using the \textit{following the leader} \cite{dasgupta_-line_2007} approach described as
\begin{subequations}
\begin{align}
& \varpi_k \leftarrow \varpi_k + \frac{1}{N} \big(o_k - \varpi_k\big) \\ 
& \mu_k \leftarrow \mu_k + o_k\big( \frac{x_{new}-\mu_k}{\varpi_k N + 1} \big) \\
& \sigma_k^2 \leftarrow \sigma_k^2 + o_k \bigg(\frac{\varpi_k N (x_{new}-\mu_k)^2}{(\varpi_k N + 1)^2} -\frac{\sigma_k^2}{\varpi_k N + 1}  \bigg)
\end{align}
\label{eq:updating_equations}
\end{subequations}
where $\sigma_k^2$ is equal to $\tau_k^{-1}$.

When the new observed sample cannot be modeled by the existing components, a new component is created with mixing coefficient $\varpi_{new}$, mean value $\mu_{new}$ and standard deviation $\sigma_{new}$, the parameters of which are given as
\begin{equation}
\varpi_{new} = \frac{1}{N}\:\:,\:\:\mu_{new} = x_{new}\:\:,\:\:\sigma_{new}^2 = \frac{(2\epsilon)^2 - 1}{12}
\label{eq:new_component}
\end{equation} 

From (\ref{eq:new_component}), we see that the mixing coefficient for the new component is equal to $1/N$ since it models only one sample (the new observed one), its mean value equals the value of the new sample and its variance the variance the Uniform distribution, whose the lower and upper bounds are $x_{new}-\epsilon$ and $x_{new}+\epsilon$ respectively. When a new component is created the values for the parameters for all the other components remain unchanged except mixing coefficients $\bm \varpi$, which are normalized to sum $\frac{N-1}{ N}$. After each adaptation of the system to new observed samples, either they modeled by the trained model or not, the mixing coefficients of the components are normalized to sum to one.

Figure \ref{fig:km vs mog} presents the adaptation of our model (first row) and the model presented in \cite{zivkovic_improved_2004} (second row) to new observed data. To evaluate the quality of the adaptation of the models, we used a toy dataset with 100 observations. Observed data were generated from two Normal distributions with mean values 16 and 50 and standard deviations 1.5 and 2.0 respectively. The initially trained models are presented in the left column. Then, we generated 25 new samples form a Normal distribution with mean value 21 and standard deviation 1.0. Our model creates a new component and successfully fits the data. On the contrary, the model of \cite{zivkovic_improved_2004} is not able to capture the statistical relations of the new observations and fails to separate the data generated from distributions with mean values 16 and 21 (middle column). The quality of our adaptation mechanism becomes more clear in the right column, which presents the adaptation of both models after 50 new observations. 

\begin{tabular}{ l }
  \hline \hline                      
  \textbf{Algorithm 1}: Overview of Background Subtraction  \\
  \hline
  1:\:\:\: capture $N$ frames \\
  2:\:\:\: create $N$-length history for each pixel \\
  3:\:\:\: initialize parameters (see Section \ref{sec:online adaptation mechanism}) \\
  4:\:\:\: \textbf{until} convergence (training phase: Section \ref{sec:random variables optimization}) \\
  5:\:\:\:\:\:\:\:\:\: compute $r_{nk}$ using (\ref{eq:r_nk}) \\
  6:\:\:\:\:\:\:\:\:\: recompute parameters using (\ref{eq:q_star_varpi_derivation}), (\ref{eq:q_star_mk_tauk}) and (\ref{eq:q_star_tauk}) \\
  7:\:\:\: \textbf{for each} new captured frame \\
  8:\:\:\:\:\:\:\:\:\: classify each pixel as foreground or background \\
  \:\:\:\:\:\:\:\:\:\:\:\:\: (see Section \ref{sec:background subtraction}) \\
  9:\:\:\:\:\:\:\:\:\: update background model (see Section \ref{sec:online adaptation mechanism}) \\
  \hline  
\end{tabular}

\section{Background subtraction}
\label{sec:background subtraction}

In this section we utilize our model for background subtraction. We initially capture $N$ frames used to create an infrared responses history for each pixel. These histories act as observed data and used to train a model for each pixel. To classify a pixel of a new captured frame as background or foreground, we compute the probability its value to be represented by the mixture model. If this value is larger than a threshold the pixel is classified as background, otherwise it is classified as foreground. The threshold can be defined in relation to the parameters of the mixture, in order to be dynamically adapted. For example, we can define the threshold to be equal to $\mu_c \pm \nu \sigma_c$ where $\nu$ is a scalar defining a confidence interval. The overview of the background subtraction algorithm is shown in Algorithm 1.

\section{Experimental results}
\label{sec:experimental results}

For evaluating our algorithm, we used the Ohio State University (OSU) thermal datasets and a dataset captured at Athens International Airport (AIA) during Evacuate\footnote{\url{http://www.evacuate.eu/}} European funding project. OSU datasets contain frames that have been captured using a thermal camera and have been converted to grayscale images. In contrast, the AIA dataset contains raw thermal frames whose pixel values correspond to the real temperature of objects.  

\begin{figure}[b]
\begin{minipage}[b]{0.33\linewidth}
  \centering
  \centerline{\includegraphics[width=5.8cm]{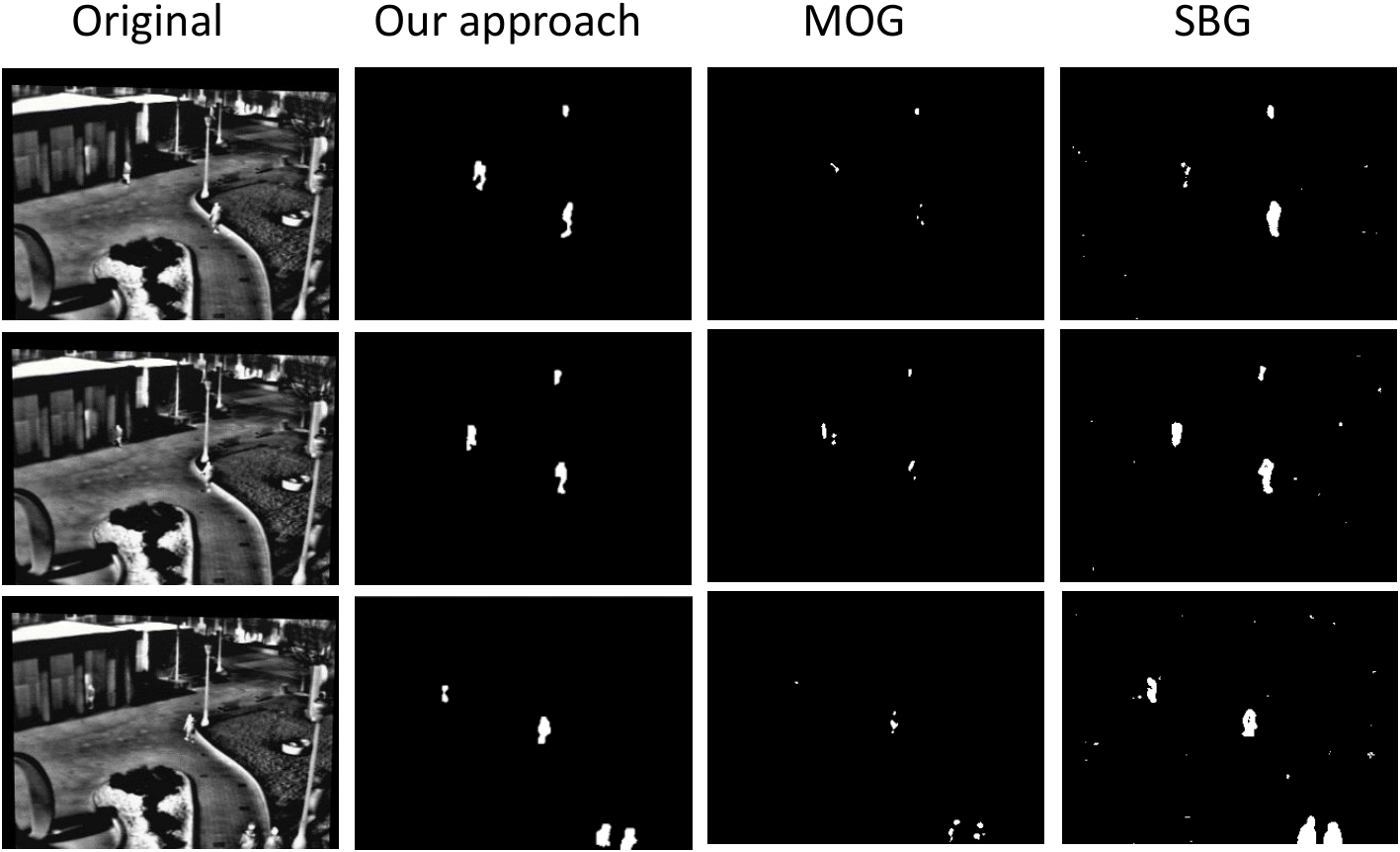}}
  \centerline{(a) OSU 1 dataset} \medskip
\end{minipage} 
\begin{minipage}[b]{0.33\linewidth}
  \centering
  \centerline{\includegraphics[width=5.8cm]{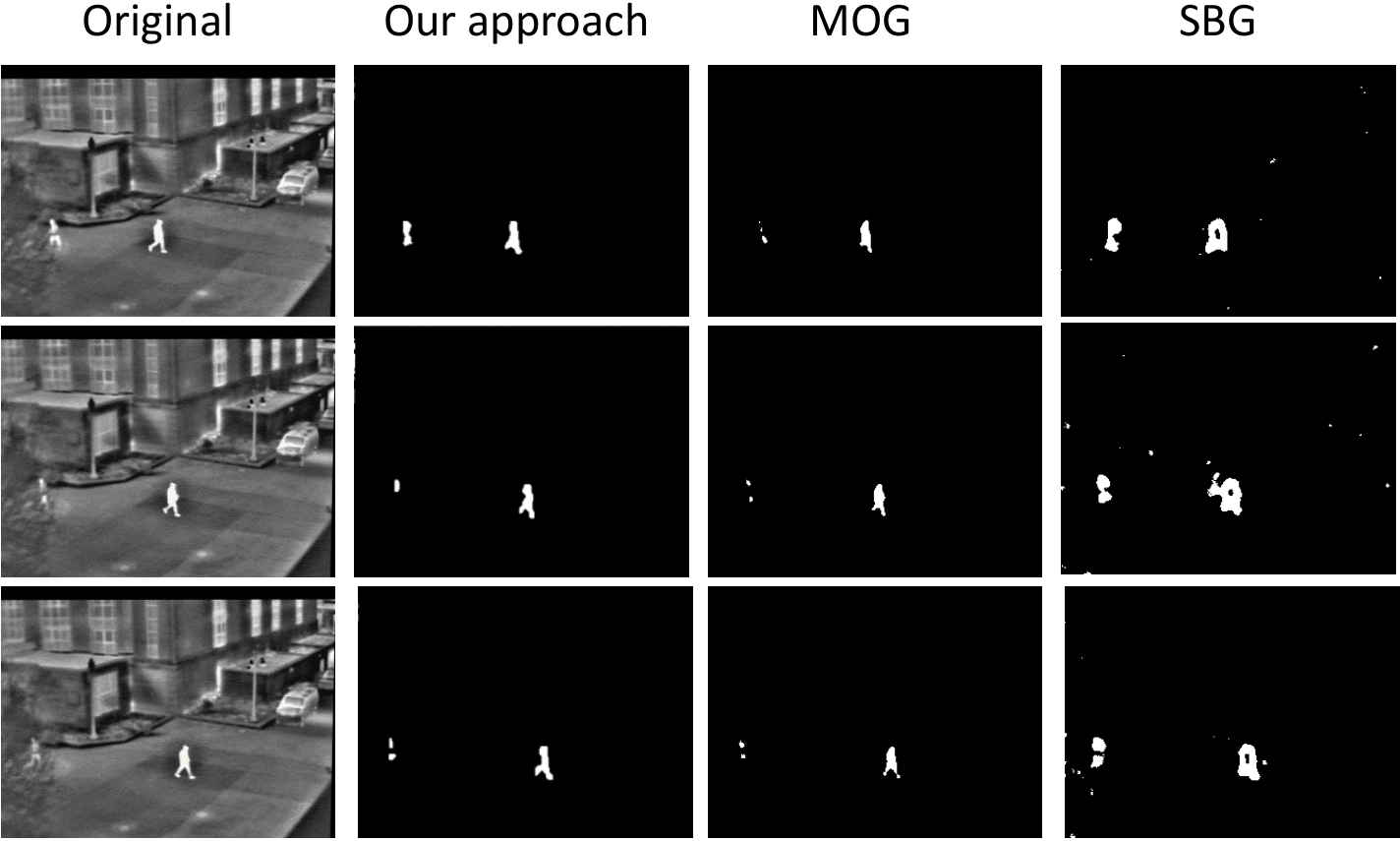}}
  \centerline{(a) OSU 2 dataset} \medskip
\end{minipage}
\begin{minipage}[b]{0.33\linewidth}
  \centering
  \centerline{\includegraphics[width=5.8cm]{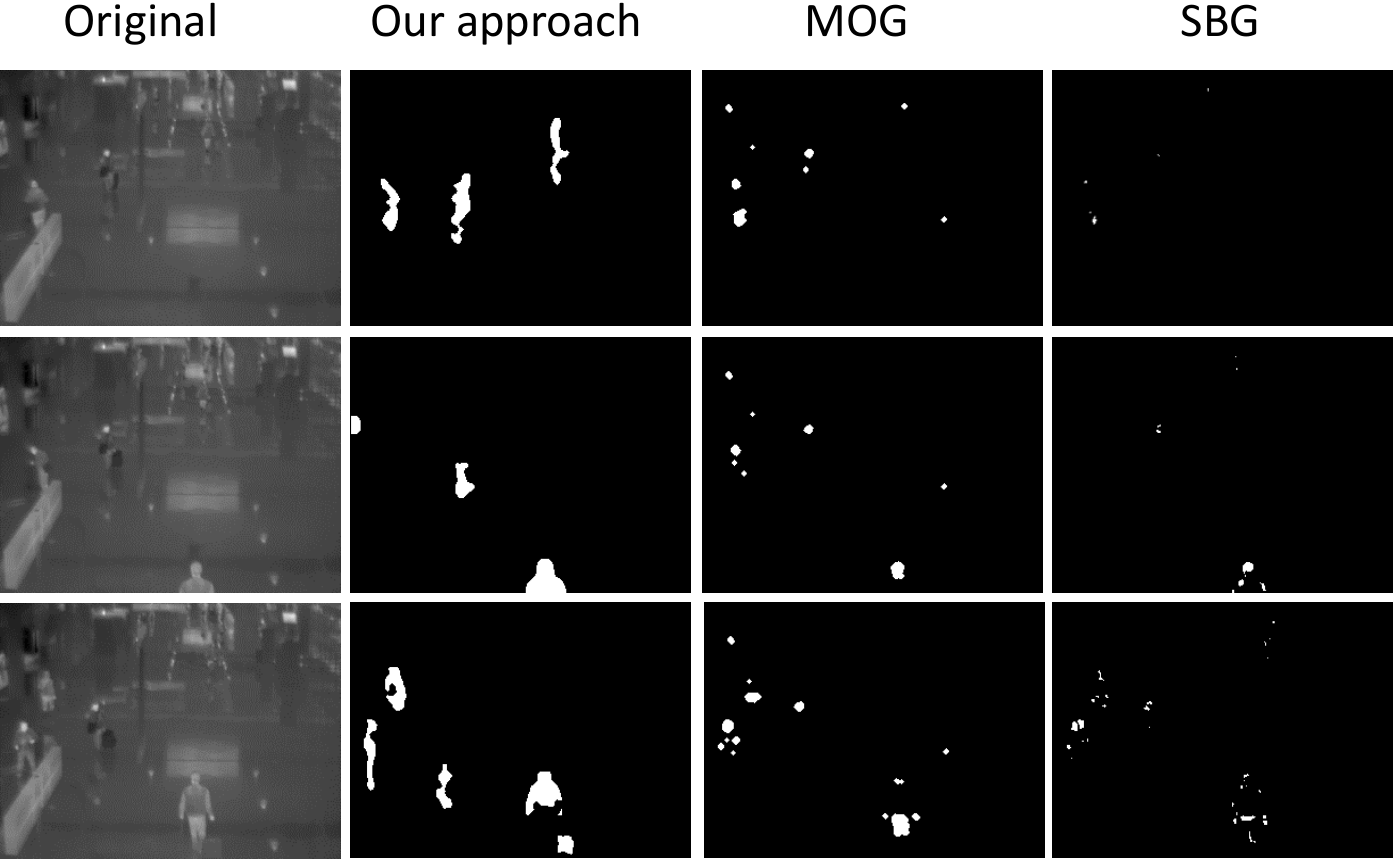}}
  \centerline{(a) AIA dataset} \medskip
\end{minipage}

\caption{Visual results for all datasets.}
\label{fig:OSU4}
\end{figure}

OSU datasets \cite{davis_fusion-based_2005, davis_robust_2004, davis_background-subtraction_2007} are widely used for benchmarking algorithms for pedestrian detection and tracking in infrared imagery. Videos were captured under different illumination and weather conditions. AIA dataset was captured using a Flir A315 camera at different Airside Corridors and the Departure Level. Totally, 10 video sequences were captured, with frame dimensions $320 \times 240$ pixels of total duration 32051 frames, at 7.5fps, that is, about 1h and 12mins. During experimentation process we used the sequence captured at the Departure Level Entrance 3 which provides a panoramic view of the space. The other sequences at corridors, due to narrow space perspective and the fact that videos were captured by mounting the camera at human height level, are inappropriate for testing background subtraction algorithms.    

\begin{figure}[t]
\begin{minipage}[b]{0.4\linewidth}
  \centering
  \centerline{\includegraphics[width=6.9cm]{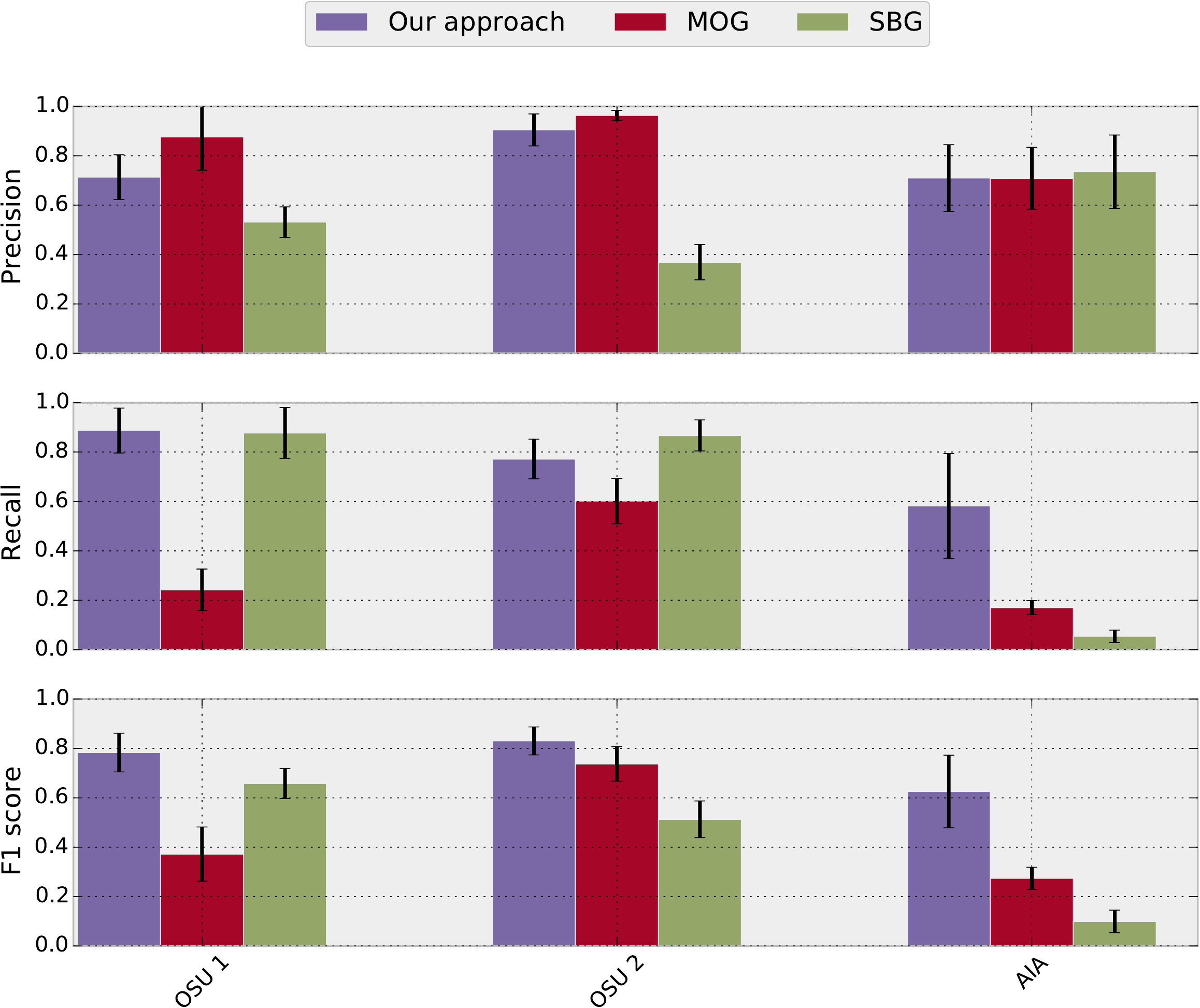}}
  \centerline{(a) Precision, recall and F1 score} \medskip
\end{minipage}
\begin{minipage}[b]{0.6\linewidth}
  \centering
  \centerline{\includegraphics[width=10.50cm]{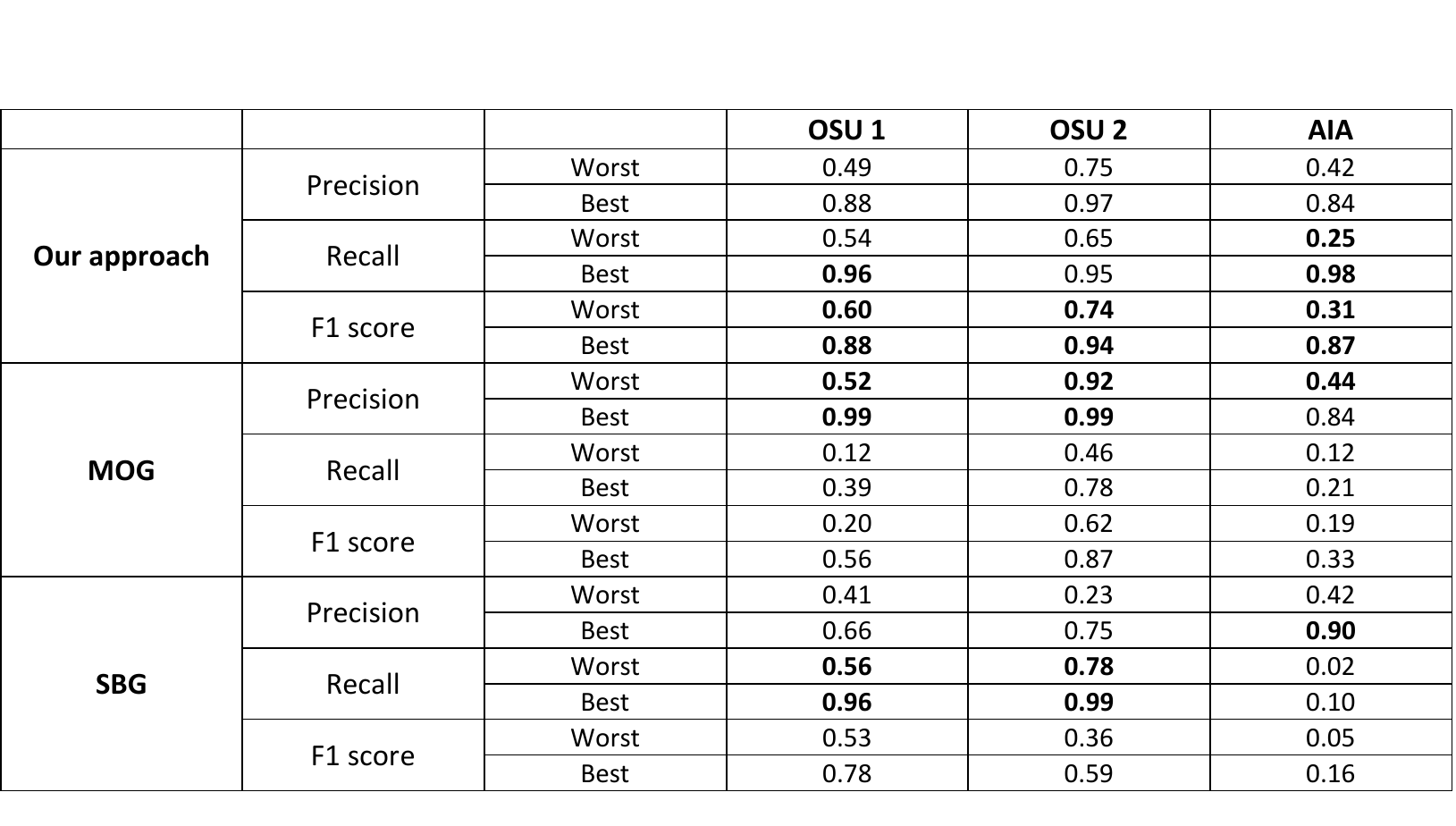}}
  \centerline{(b) Best/worst case for precision, recall and F1 score} \medskip
\end{minipage}
\caption{Algorithms performance per dataset.}
\label{fig:prf}
\end{figure}

We compared our method with the method presented by Zivkovic in \cite{zivkovic_improved_2004} (MOG), which is one of the most robust and widely used background subtraction technique, and with the method for extracting the regions of interest presented in \cite{davis_robust_2004, davis_background-subtraction_2007} (SBG) used for thermal data. To conduct the comparison we utilized the objective metrics of \textit{recall}, \textit{precision} and \textit{F1 score} on a pixel wise manner. Figures \ref{fig:OSU4} visually present the performance of the three methods. As is observed, our method outperforms both MOG and SBG on all datasets. While MOG and SBG perform satisfactory on grayscale frames of OSU datasets, their performance collapses when they applied on AIA dataset, which contains actual thermal responses. Regarding OSU datasets, MOG algorithm while presents high precision it yields very low recall values, i.e. the pixels that have been classified as foreground are indeed belong to the foreground class, but a lot of pixels that in fact belong to background have been misclassified. SBG algorithm seems to suffer by the opposite problem. Regarding AIA dataset, our method significantly outperforms both methods. In particular, while MOG and SBG algorithms present relative high precision, their recall values are under 0.2. Figure \ref{fig:prf}(a) presents average precision, recall and F1 score per dataset and per algorithm for all frames examined to give an objective evaluation. In Figure \ref{fig:prf}(b) presents the best and worst case in terms of precision, recall and F1 score among all frames examined.

Regarding computational cost, the main load of our algorithm is in the implementation of EM optimization. In all experiments conducted, the EM optimization converges within 10 iterations. Practically, the time required to apply our method is similar to the time requirements of Zivkovic's method making it suitable for real-time applications.

\section{Conclusions}
\label{sec:conclusions}

This paper presents a background subtraction method applicable to thermal imagery, based on Gaussian mixture modeling with unknown number. We analytically derive the solutions that describe the parameters of the model and we use the EM optimization to estimate their values, avoid sampling algorithms and high computational cost. Due to its low computational cost and the real-time operation, our method is suitable for real-world applications.

\appendix
\section{Appendix}
\label{ap:appendix}
Using (\ref{eq:q_star}) and (\ref{eq:joint_factorization}) the logarithm of $q^*(\bm Z)$ is given by
\begin{equation}
\begin{aligned}
\ln q^*(\bm Z) = & \mathbb{E}_{\bm \varpi}[\ln p(\bm Z|\bm \varpi)] + \\ & + \mathbb{E}_{\bm \mu, \bm \tau}[\ln p(\bm X|\bm Z, \bm \mu, \bm \tau)] + \mathcal{C}
\end{aligned}
\label{eq:q_Z_optimized_2}
\end{equation}
substituting (\ref{eq:p_Z}) and (\ref{eq:p_X}) into (\ref{eq:q_Z_optimized_2}) we get
\begin{subequations}
\begin{align}
& \ln q^*(\bm Z)=  \sum_{n=1}^{N} \sum_{k=1}^{K} z_{nk} \bigg( \mathbb{E}\big[\ln \varpi_k \big] + \frac{1}{2}\mathbb{E}\big[\ln \tau_k\big] -\nonumber \\ 
&\:\:\:\:\:\:\:\:\:\:-\frac{1}{2}\ln2\pi - \frac{1}{2} \mathbb{E}_{\bm \mu, \bm \tau}\big[(x_n-\mu_k)^2\tau_k \big]\bigg) + \mathcal{C} \Rightarrow \nonumber  
\end{align}
\label{eq:q_star_Z_derivation}
\end{subequations}  


\noindent Using (\ref{eq:joint_factorization}) and (\ref{eq:q_star}) the logarithm of $q^*(\bm \varpi, \bm \mu, \bm \tau)$ is
\begin{subequations}
\begin{align}
\ln q^*(\bm \varpi, \bm \mu, \bm \tau) & =  \mathbb{E}_{\bm Z}\big[\ln p(\bm X|\bm Z, \bm \mu, \bm \tau) + \nonumber \\ 
& + \ln p(\bm Z|\bm \varpi) + \nonumber \\ 
& + \ln p(\bm \varpi) + \ln p(\bm \mu, \bm \tau)\big] + \mathcal{C} = \\
& = \sum_{n=1}^{N}\sum_{k=1}^{K} \mathbb{E}\big[z_{nk}\big] \ln \mathcal{N}(x_n|\mu_k, \tau_k^{-1}) + \nonumber \\ 
&+ \mathbb{E}_{\bm Z}\big[\ln p(\bm Z|\bm \varpi)\big] \nonumber \\
& + \ln p(\bm \varpi) + \sum_{k=1}^{K}\ln p(\mu_k, \tau_k) + \mathcal{C}
\label{eq:q_varpi_mu_tau_optimized_b}
\end{align}
\label{eq:q_varpi_mu_tau_optimized}
\end{subequations} 
Due to the fact that there is no term in (\ref{eq:q_varpi_mu_tau_optimized_b}) that contains parameters from both sets $\{\bm \varpi\}$ and $\{\bm \mu, \bm \tau\}$, the distribution $q^*(\bm \varpi, \bm \mu, \bm \tau)$ can be factorized as $q(\bm \varpi, \bm \mu, \bm \tau) = q(\bm \varpi) \prod_{k=1}^{K}q(\mu_k, \tau_k)$.
The distribution for $q^*(\bm \varpi)$ is derived using only those terms of (\ref{eq:q_varpi_mu_tau_optimized_b}) that depend on the variable $\bm \varpi$. Therefore the logarithm of $q(\bm \varpi)$ is given by
\begin{subequations}
\begin{align}
\ln q^*(\bm \varpi) & = \mathbb{E}_{\bm Z}\big[\ln p(\bm Z|\bm \varpi)\big] + \ln p(\bm \varpi) + \mathcal{C} = \\
& = \sum_{k=1}^{K} \ln \varpi_k^{(\sum_{n=1}^{N}r_{nk} + \lambda_0 -1)} + \mathcal{C} = \\
& = \sum_{k=1}^{K} \ln \varpi_k^{(N_k + \lambda_0 -1)} + \mathcal{C}
\label{eq:q_star_varpi_derivation_c}
\end{align}
\end{subequations}  
We have made use of $\mathbb{E}[z_{nk}]=r_{nk}$, and we have denote as $N_k=\sum_{n=1}^{N}r_{nk}$. (\ref{eq:q_star_varpi_derivation_c}) suggests that $q^*(\bm \varpi)$ is a Dirichlet distribution with hyperparameters $\bm \lambda = \{N_k + \lambda_0\}_{k=1}^K$.

Using only those terms of (\ref{eq:q_varpi_mu_tau_optimized_b}) that depend on variables $\bm \mu$ and $\bm \tau$, the logarithm of $q^*(\mu_k, \tau_k)$ is given by

\begin{align}
\ln q^*(\mu_k, \tau_k) & = \ln \mathcal{N}(\mu_k|m_0, (\beta_0 \tau_k)^1) + \nonumber \\ 
&\:\:\:\:\: + \ln Gam(\tau_k|a_0, b_0) + \nonumber \\
&\:\:\:\:\: + \sum_{n=1}^{N}\mathbb{E}\big[z_{nk}\big]\ln \mathcal{N}(x_n|\mu_k,\tau_k^{-1}) + \mathcal{C} = \nonumber \\ 
& = -\frac{\beta_0\tau_k}{2}(\mu_k-m_0)^2 + \frac{1}{2}\ln (\beta_0 \tau_k) + \nonumber \\ 
&\:\:\:\:\: + (a_0-1)\ln \tau_k - b_0\tau_k - \nonumber \\ &\:\:\:\:\: -\frac{1}{2}\sum_{n=1}^{N}\mathbb{E}\big[z_{nk}\big](x_n-\mu_k)^2\tau_k + \nonumber \\ 
&\:\:\:\:\: + \frac{1}{2}\bigg(\sum_{n=1}^{N}\mathbb{E}\big[z_{nk}\big] \bigg) \ln(\beta_0\tau_k) + \mathcal{C}
\label{eq:q_star_mu_tau_derivation_b}
\end{align} 
For the estimation of $q^*(\mu_k|\tau_k)$, we use (\ref{eq:q_star_mu_tau_derivation_b}) and keep only those factors that depend on $\mu_k$.
\begin{subequations}
\begin{align}
\ln q^*(\mu_k|\tau_k) & = -\frac{\beta_0\tau_k}{2}\big(\mu_k-m_0\big)^2 - \nonumber \\
&\:\:\:\:\: - \frac{1}{2}\sum_{n=1}^{N}\mathbb{E}\big[z_{nk}\big]\big(x_n - \mu_k\big)^2\tau_k = \\
& = -\frac{1}{2}\mu_k^2\Big(\beta_0 + N_k\Big)\tau_k +\nonumber \\ 
&\:\:\:\:\: + \mu_k \tau_k \Big(\beta_0 m_0 + N_k\bar x_k\Big) + \mathcal{C} \Rightarrow \\
& q^*(\mu_k|\tau_k) = \mathcal{N}(\mu_k|m_k, (\beta_k \tau)^{-1})
\label{eq:q_star_mu_derivation_b}
\end{align}
\label{eq:q_star_mu_derivation}
\end{subequations}  
where $\bar x_k = \frac{1}{N_k}\sum_{n=1}^{N}r_{nk}x_n$, $\beta_k = \beta_0 + N_k$ and $m_k = \frac{1}{\beta_k}(\beta_0 m_0 + N_k \bar x_k)$. 

After the estimation of $q^*(\mu_k|\tau_k)$, logarithm of the optimized the distribution $q^*(\tau_k)$ is given by
\begin{subequations}
\begin{align}
\ln q^*(\tau_k) & = \ln q^*(\mu_k, \tau_k) - \ln q^*(\mu_k|\tau_k) = \\
& = \bigg(a_0+\frac{N_k}{2} - 1\bigg) \ln \tau_k - \nonumber \\ &\:\:\:\:\:\: -\frac{1}{2}\tau_k\bigg(\beta_0\big(\mu_k-m_0\big)^2 + \nonumber \\
&\:\:\:\:\:\: +2b_0 + \sum_{n=1}^{N}r_{nk}\big(x_n-\mu_k\big)^2 -\nonumber \\ &\:\:\:\:\:\: -\beta_k\big(\mu_k-m_k\big)^2 \bigg) + \mathcal{C} \Rightarrow \\
& q^*(\tau_k) = Gam(\tau_k|a_k, b_k)
\label{eq:q_star_tauk_derivation_b}
\end{align}
\label{eq:q_star_tauk_derivation}
\end{subequations}
The parameters $a_k$ and $b_k$ are given by 
\begin{subequations}
\begin{align}
a_k & = a_0 +  \frac{N_k}{2} \\
b_k & = b_0 + \frac{1}{2}\bigg(N_k\sigma_k + \frac{\beta_0 N_k}{\beta_0 + N_k}\big(\bar x_k - m_0\big)^2 \bigg)
\end{align}
\label{eq:ap_q_star_tauk}
\end{subequations}
where $\sigma_k = \frac{1}{N_k}\sum_{n=1}^{N}(x_n-\bar x_k)^2$.

{\small
\bibliographystyle{ieee}
\bibliography{Arxiv_makantasis}
}

\end{document}